%% file: paper.tex
\documentclass[letterpaper]{article}
\usepackage{aaai}
\usepackage{times}
\usepackage{helvet}
\usepackage{courier}
\usepackage{url}
\usepackage{graphicx}
\usepackage[utf8]{inputenc} 
\usepackage{url}            
\usepackage{booktabs}       
\usepackage{amsfonts}       
\usepackage{nicefrac}       

\usepackage{subcaption}
\usepackage{tabularx}
\usepackage{booktabs}
\usepackage{color}
\usepackage{caption}
\usepackage{subcaption}
\usepackage{amsmath, amsthm, amssymb}

\title{Meta-descent for Online, Continual Prediction}

\author{
Andrew Jacobsen,\textsuperscript{1}
Matthew Schlegel,\textsuperscript{1}
Cameron Linke,\textsuperscript{1}\\
{\bf \Large Thomas Degris,\textsuperscript{2}
Adam White, \textsuperscript{1,3}
Martha White \textsuperscript{1}}\\
\textsuperscript{1}University of Alberta, Edmonton, Canada,\\
\textsuperscript{2}Google DeepMind, London, UK\\
\textsuperscript{3}Google DeepMind, Edmonton, Canada\\
ajjacobs@ualberta.ca, mkschleg@ualberta.ca, clinke@ualberta.ca\\
thomas.degris@gmail.com, amw8@ualberta.ca, whitem@ualberta.ca
}

\newcommand{\citet}[1]{\citeauthor{#1} ~\shortcite{#1}}
\newcommand{\citep}{\cite}

\newcommand{\Gmat}{\mathbf{G}}
\newcommand{\Hmat}{\mathbf{H}}

\newcommand{\defeq}{\mathrel{\overset{\makebox[0pt]{\mbox{\normalfont\tiny\sffamily def}}}{=}}}
\newcommand{\elwise}{\circ}
\newcommand{\veczeros}[1]{
\bigg[\!\! \begin{array}{c}
\zerovec\\
#1\\
\zerovec
\end{array} \!\!\bigg]}
\newcommand{\diag}{\text{diag}}

\newcommand{\loss}{\ell}

\newcommand{\Psimat}{\boldsymbol{\Psi}}

\newcommand{\dvec}{\mathbf{d}}
\newcommand{\evec}{\mathbf{e}}
\newcommand{\gvec}{\mathbf{g}}
\newcommand{\hvec}{\mathbf{h}}

\newcommand{\ovec}{\mathbf{o}}

\newcommand{\uvec}{\mathbf{u}}

\newcommand{\wvec}{\mathbf{w}}

\newcommand{\zerovec}{\mathbf{0}}

\newcommand{\epsilonvec}{\boldsymbol{\epsilon}}

\newcommand{\psivec}{\boldsymbol{\psi}}

\newcommand{\xdim}{d}

\newcommand{\nparams}{k}

\newcommand{\stepsize}{\alpha}

\newcommand{\metastep}{\bar{\stepsize}}
\newcommand{\stepsizevec}{\boldsymbol{\alpha}}

\newcommand{\betavec}{\boldsymbol{\beta}}

\newcommand{\featdim}{d}

\newcommand{\inv}{{-1}}

\newcommand{\eye}{\mathbf{I}}

\newcommand{\E}{\mathbb{E}}
\newcommand{\RR}{\mathbb{R}}
\newcommand{\sign}{\mathrm{sign}}

\newcommand{\lowerthreshold}{\epsilon}

\newcommand{\featvec}{\mathbf{x}}
\newcommand{\weights}{\wvec}

\newcommand{\Hess}{\mathbf{H}}

\newcommand{\myparagraph}[1]{\textbf{#1}}

\begin{document}
\maketitle
\begin{abstract}
This paper investigates different vector step-size adaptation approaches for non-stationary online, continual prediction problems. Vanilla stochastic gradient descent can be considerably improved by scaling the update with a vector of appropriately chosen step-sizes. Many methods, including AdaGrad, RMSProp, and AMSGrad, keep statistics about the learning process to approximate a second order update---a vector approximation of the inverse Hessian. Another family of approaches use meta-gradient descent to adapt the step-size parameters to minimize prediction error. These meta-descent strategies are promising for non-stationary problems, but have not been as extensively explored as quasi-second order methods. We first derive a general, incremental meta-descent algorithm, called AdaGain, designed to be applicable to a much broader range of algorithms, including those with semi-gradient updates or even those with accelerations, such as RMSProp. We provide an empirical comparison of methods from both families. We conclude that methods from both families can perform well, but in non-stationary prediction problems the meta-descent methods exhibit advantages. Our method is particularly robust across several prediction problems, and is competitive with the state-of-the-art method on a large-scale, time-series prediction problem on real data from a mobile robot.
\end{abstract}

\section{Introduction}
In this paper we consider continual, non-stationary prediction problems. Consider a learning system whose objective is to learn a large collection of predictions about an agent's future interactions with the world. The predictions specify the value of some signal many steps in the future, given that the agent follows some specific course of action. There are many examples of such prediction learning systems including Predictive State Representations \citep{littman2001predictive}, Observable Operator Models \citep{jaeger2000observable}, Temporal-difference Networks \citep{sutton2004temporal}, and General Value Functions \citep{sutton2011horde}. In our setting, the agent continually interacts with the world, making new predictions about the future, and revising its previous predictions as new outcomes are revealed. Occasionally, partially due to changes in the world and partially due to changes in the agent's own behaviour, the targets may change and the agent must refine its predictions.  
\footnote{We exclude recent meta-learning frameworks (MAML \citep{finn2017model}, LTLGDGD \citep{andrychowicz2016learning}) because they assume access to a collection of tasks that can be sampled independently, enabling the agent to learn how to select meta-parameters for a new problem. In our setting, the agent must solve a large collection of non-stationary prediction problems in parallel using off-policy learning methods. 
} 

Stochastic gradient descent (SGD) is a natural choice for our setting because gradient descent methods work well when paired with abundant training data. The performance of SGD is dependent on the step-size parameter (scalar, vector or matrix), which scales the gradient to mitigate sample variance and improve data efficiency. Most modern large-scale learning systems make use of optimization algorithms that attempt to approximate stochastic second-order gradient descent to adjust both the direction and magnitude of the descent direction, with early work indicating the benefits of such quasi-second order methods if used carefully in the stochastic case \citep{schraudolph2007astochastic,bordes2009sgd}. 
Many of these algorithms attempt to approximate the diagonal of the inverse Hessian, which describes the curvature of the loss function, and so maintain a vector of step-sizes---one for each parameter. Starting from AdaGrad \citep{mcmahan2010adaptive,duchi2011adaptive}, several diagonal approximations have been proposed, including RmsProp \citep{tielman2012rmsprop}, AdaDelta \citep{zeiler2012adadelta}, vSGD \citep{schaul2013no}, Adam \citep{kingma2015adam} and AmsGrad \citep{reddi2018onthe}.
Stochastic quasi-second order updates have been derived specifically for
temporal difference learning, with some empirical success \citep{meyer2014accelerated}, particularly in terms of parameter sensitivity \citep{pan2017accelerated,pan2017effective}. On the other hand, second order methods, by design, assume the loss and thus Hessian are fixed, and so non-stationary dynamics or drifting targets could be problematic. 

A related family of optimization algorithms, called \emph{meta-descent} algorithms, were developed for continual, online prediction problems. These algorithms perform meta-gradient descent adapting a vector of step-size parameters to minimize the error of the base learner, instead of approximating the Hessian.
Meta-descent applied to the step-size was first introduced for online least-mean squares methods \citep{jacobs1988increased,sutton1992adapting,sutton1992gain,almeida1999parameter,mahmood2012tuning}, including the linear complexity method IDBD \citep{sutton1992adapting}. IDBD was later extended to more general losses \citep{schraudolph1999local} and to support (semi-gradient) temporal difference methods \citep{dabney2012adaptive,dabney2014thesis,kearney2018tidbd}. 
These methods are well-suited to non-stationary problems, and have been shown to ignore irrelevant features.  The main limitation of several of these meta-descent algorithms, however, is that the derivations are heuristic, making it difficult to extend to new settings beyond linear temporal difference learning. The more general approaches, like Stochastic Meta-Descent (SMD) \citep{schraudolph1999local}, require the update to be a stochastic gradient descent update and have some issues in biasing towards smaller step-sizes \citep{wu2018understanding}. It remains an open challenge to make these meta-descent strategies as broadly and easily applicable as the AdaGrad variants. 
In this paper we introduce a new meta-descent algorithm, called AdaGain, that
attempts to optimize the stability of the base learner, rather than convergence
to a fixed point. AdaGain is built on a generic derivation scheme that allows it
to be easily combined with a variety of base-learners including SGD, (semi-gradient) temporal-difference learning and even optimized SGD updates, like AMSGrad. 
%
%
Our goal is to investigate the utility of both meta-descent methods and the more widely used quasi-second order optimizers in online, continual prediction problems.
We provide an extensive empirical comparison on (1) canonical optimization
problems that are difficult to optimize with large flat regions (2) an online,
supervised tracking problem where the optimal step-sizes can be computed, (3) a
finite Markov Decision Process with linear features that cause conventional
temporal difference learning to diverge, and (4) a high-dimensional time-series
prediction problem using data generated from a real mobile robot. In problems
with non-stationary dynamics the meta-descent methods can exhibit an advantage
over the quasi-second order methods. On the difficult optimization problems,
however, meta-descent methods fail, which, retrospectively, is unsurprising
given the meta-optimization problem for stepsizes is similarly difficult to
optimize.
We show that AdaGain can possess the advantages of both families --- performing
well on both optimization problems with flat regions as well as non-stationary
problems --- by selecting an appropriate base learner, such as RMSProp.


%
%
%

\section{Background and Notation}
In this paper we consider {\em online continual prediction} problems modeled as non-stationary, uncontrolled dynamical systems. On each discrete time step $t$, the agent observes the internal state of the system through an imperfect summary vector $\ovec_t \in \mathcal{O} \in \mathbb{R}^d$ for some $d\in\mathbb{N}$, such as the sensor readings of a mobile robot. 
On each step, the agent makes a prediction about a target signal $T_t \in \mathbb{R}$. 
In the simplest case, the target of the prediction is a component $i$ of the observation vector on the next step $T_t  = \ovec_{t+1,i}$---the classic one-step prediction. In the more general case, the target is constructed by mapping the entire future of the observation time series to a scalar, such as
the discounted sum formulation used in reinforcement learning: $T_t = \mathbb{E}[\sum_{k=0}^\infty \gamma^{k}\ovec_{t+ k+1,i}]$, where $\gamma\in [0, 1)$ discounts the contribution of future observations to the infinite sum.
The prediction $P_t \in \RR$ is generated by a parametrized function, with modifiable parameter vector $\wvec_t \in \RR^\nparams$.


In online continual prediction problems the agent updates its predictions (via $\wvec_t$) with each new sample $\ovec_t$, unlike the more common batch and stochastic settings. 
The agent's objective is to minimize the error between the prediction $P_t$ given by $\wvec_t$ and the target $T_t$ before it is observed, over all time steps.
Online continual prediction problems are typically solved using stochastic updates to adapt the parameter vector $\wvec_t$ after each time step $t$ to reduce the error (retroactively) between $P_t$ and $T_t$. 
Generically, for stochastic {\em update vector} $\Delta_t \in \RR^\xdim$, the weights are modified
\begin{equation}\label{eq_generic}
\wvec_{t+1} = \wvec_t + \stepsizevec_t \circ \Delta_t
\end{equation}
for a vector step-size $\stepsizevec_t$, where the operator $\elwise$ denotes element-wise multiplication. 
Given an {update vector}, the goal is to select $\stepsizevec_t$ to reduce error, into the future. Semi-gradient methods like temporal difference learning follow a similar scheme, but $\Delta_t$ is not the gradient of an objective function. 

Step-size adaptation for the stationary setting is often based on estimating
second-order updates.\footnote{A related class of algorithms are natural
  gradient methods, which aim to be robust to the functional parametrization.
  Incremental natural gradient methods have been proposed
  \citep{amari2000adaptive}, including for policy evaluation with gradient TD
  methods \citep{dabney2014natural}. However, these algorithms do not remove the
  need select a step-size, and so we do not consider them further here.} The
idea is to estimate the loss function $\loss: \RR^\xdim \rightarrow \RR$ locally
around the current weights $\wvec_t$ using a second-order Taylor series approximation---which requires the Hessian $\Hmat_t$. A closed-form solution can then be obtained for the approximation, because it is a quadratic function, giving the next candidate solution $\wvec_{t+1} = \wvec_t - \left(\Hess_{t}\right)^\inv\nabla \loss(\wvec_t)$. 
%
If instead the Hessian is approximated---such as with a diagonal approximation---then we obtain \emph{quasi-second order} updates.
Taken to the extreme, with the Hessian approximated by a scalar, as $ \Hess_{t} = \stepsize_t^\inv \eye$, we obtain first-order gradient descent with a step-size of $\alpha_t$. For the batch setting, the gains from second order methods are clear, with a convergence rate\footnote{There is a large literature on accelerated first-order descent methods, starting from early work on momentum \citep{nesterov1983amethod} and many since focused mainly on variance reduction (c.f. \citep{roux2012astochastic}). These methods can complement step-size adaptation, but are not well-suited to non-stationary problems because many of the algorithms are designed for a batch of data and focus on increasing convergence rate to a fixed minimum.} of $O(1/t^2)$, as opposed to $O(1/t)$ for first-order descent. 

These gains are not as clear in the stochastic setting, but diagonal approximations appear to provide an effective balance between computation and convergence rate improvements \citep{bordes2009sgd}. \citet{duchi2011adaptive} provide a general regret analysis for diagonal approximations methods proving sublinear regret if step-sizes decrease to zero overtime. One algorithm, AdaGrad, uses the vector step-size $\stepsizevec_t = \eta (\sum_{i=1}^t \Delta_t + \epsilon)^\inv$ for a fixed $\eta > 0$ and a small $\epsilon > 0$, with element-wise division. RMSProp and Adam---which are not guaranteed to obtain sublinear regret---use a running average rather than a sum of gradients, with Adam additionally including a momentum term for faster convergence. AMSGrad is a modification of Adam, that satisfies the regret criteria, without decaying the step-sizes as aggressively as AdaGrad.  

The \emph{meta-descent} strategies instead directly learn step-sizes that minimize the same objective as the base learner. A simpler set of such methods, called \emph{hypergradient} methods \citep{jacobs1988increased,almeida1999parameter,baydin2018online}, only adjust the step-size based on its impact on the weights on a single step. Hypergradient Descent (HD) \citep{baydin2018online} takes the gradient of the loss $\loss(\weights)$ w.r.t. a scalar step-size $\stepsize > 0$, to get the meta-gradient for the step-size as $\partial \loss(\weights_t)/\partial \stepsize = - \nabla_\weights \loss(\weights_{t-1})^\top \nabla_\weights \loss(\weights_{t})$. The update simply requires storing the vector $\gvec_{t-1} = \nabla_\weights \loss(\weights_{t-1})$ and updating $\stepsize_{t+1} = \stepsize_t + \metastep \gvec_{t-1}^\top \gvec_{t}$, for a meta step-size $\metastep > 0$. 
More generally, meta-descent methods, like IDBD \citep{sutton1992adapting} and SMD \citep{schraudolph1999local}, consider the impact of the step-size back in time, through the weights, with $w_{t,j}$ the $j$-th element in vector $\wvec_t$
\begin{equation}
\frac{\partial \loss(\weights_t(\stepsizevec)) }{\partial \stepsize_i}
= \sum_{j}^\nparams \frac{\partial \loss(\weights_t(\stepsizevec))}{\partial w_{t,j}} \frac{\partial w_{t,j}}{\partial \stepsize_i} \label{eq_obj_smd}
.
\end{equation}
The goal is to approximate this gradient efficiently, usually using a recursive
strategy. We derive such a strategy for AdaGain below using a different meta-descent objective, and for completeness include the derivation for the SMD objective in the appendix (as the original contains an error).

\subsection{Illustrative example}
\begin{figure}[h!]
  \centering
  \includegraphics[width=0.3\textwidth]{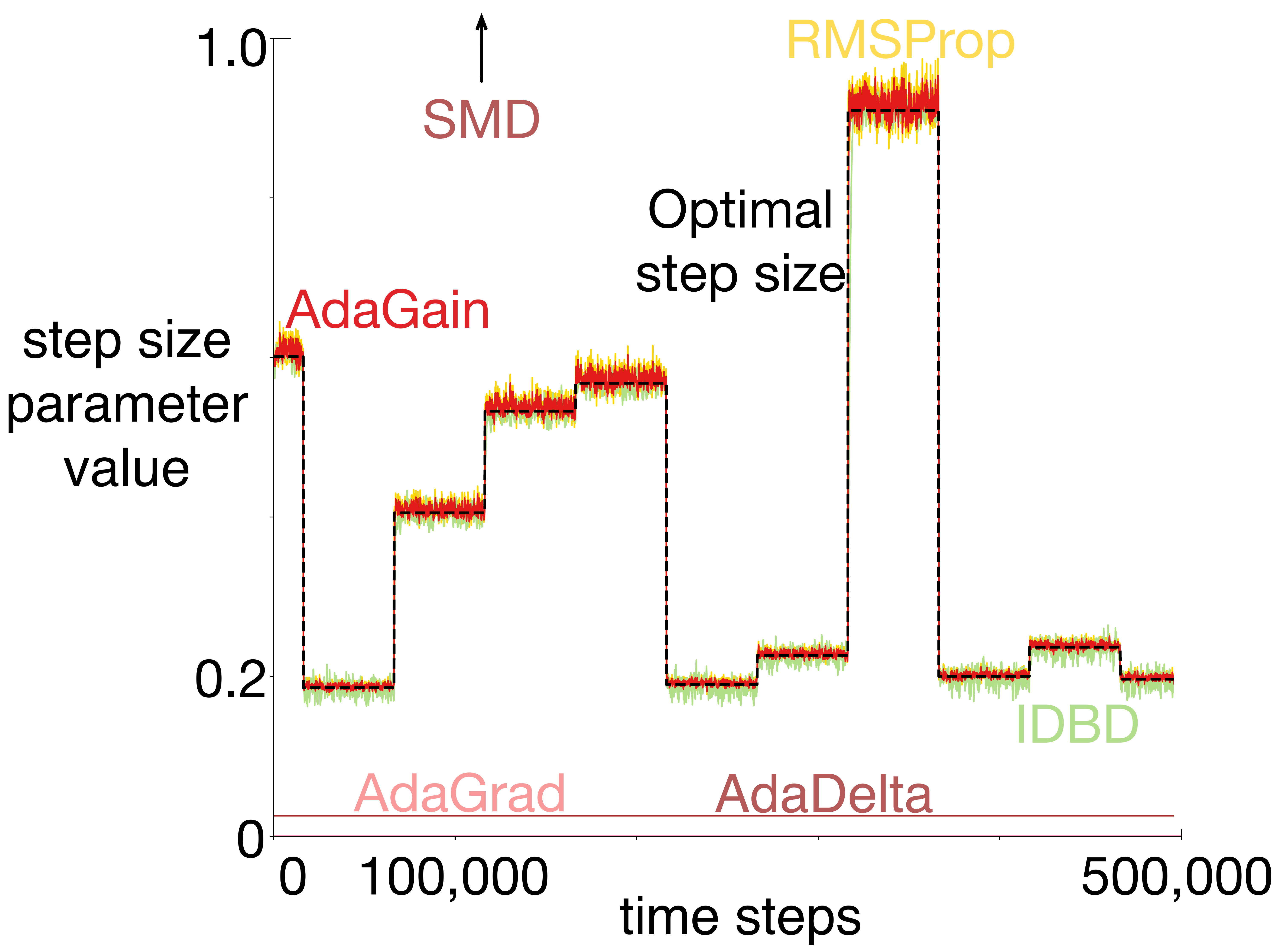}
  \caption{Optimal Gain Experiment. Depicted is the last 500,000 steps
  out of $3*(10^9)$. AdaGrad, and AdaDelta fail to learn the correct progression of stepsizes, and SMD diverges.}\label{fig:gain_intro}
\end{figure}

To make the problem more concrete, consider a simple state-less tracking problem driven by two interacting Gaussians:
{\small
\begin{equation}Y_t \defeq Z_t + \mathcal{N}(0, \sigma^2_{Y,t}), \hspace{.25cm}
Z_{t+1} \leftarrow Z_{t} + \mathcal{N}(0,\sigma^2_{Z,t}).\label{drift}
\end{equation}
}
where the agent only observes the sequence $Y_1, Y_2, \ldots$. 
The objective is minimize mean squared error (MSE) between a scalar prediction $P_t  = w_t$ and the target $T_t =Y_{t+1}$. This problem is non-stationary because $\sigma_{Y,t}$ and $\sigma_{Z,t}$ change periodically and the agent has no knowledge of the schedule. Since $\sigma_{Y,t}$ and $\sigma_{Z,t}$ govern how quickly the mean $Z_t$ drifts and the sampling variance in $Y_t$, the agent must step its step-size accordingly: larger $\sigma_{Z,t}$ requires larger stepsize, larger $\sigma_{Y,t}$ requires a smaller step-size. The agent must continually change its scalar step-size value in order to achieve low MSE. The optimal constant scalar step-size can be computed in this simple domain \citep{sutton1992adapting}, and is shown by the black dashed line in Figure \ref{fig:gain_intro}.
We compared the step-sizes learned by several well-know quasi-second order methods (AdaGrad, RMSProp, Adadelta) and three meta-descent strategies including our own AdaGain. We ran the experiment for over 24 hours to test the robustness of these methods in a long-running continual prediction task. Several methods including AdaGain were able to match the optimal step-size. However, several well-known methods including AdaGrad and AdaDelta completely fail in this problem. In addition, the meta-descent strategy SMD diverged after 8183817 time steps, highlighting the special challenges of online, continual prediction problems.    

\section{Adaptive Gain for Stability}

Tracking---continually updating the weights with recent experience---contrasts the typical goal of convergence.
Much of the previous algorithm development for step-size adaptation, however, has been towards the aim of convergence, with algorithms like AdaGrad and AMSGrad that decay step-sizes over time. Assuming finite representational capacity, there may be aspects of the problem that can never be accurately modeled or predicted by the agent. In these partially observable problems tracking and thus treating the problem as if it were non-stationary can improve prediction accuracy compared with methods that converge \citep{sutton2007ontherole}. 
In continual learning we assume the agent's task partially observable in this way, and develop a new step-size method that can facilitate tracking.  

We treat the learning system as a dynamical system---where the weight update is based
on stochastic updates known to suitably track the targets---and consider the choice of step-size as the inputs to
the system to maintain \emph{stability}. Such a view has been previously considered under adaptive gain
for least-mean squares (LMS) \citep[Chapter 4]{benveniste1990adaptive}, where weights are treated as state following a random drift. 
To generalize this idea to other incremental algorithms, we propose a more general criteria based on the magnitude of the {update vector}.


A criteria for $\stepsizevec$ to maintain stability
in the system is to keep the norm of the {update vector} small
\begin{equation}\label{eq_criteria}
\min_{\stepsizevec > 0} \E \left[\| \Delta_t(\wvec_t(\stepsizevec)) \|_2^2  \ \big| \ \wvec_0 \right]
.
\end{equation}
%
%
The update $\Delta_t(\wvec_t(\stepsizevec))$ on this time step is dependent on the step-size $\stepsizevec$
because that step-size influences $\wvec_t$ and past updates. The expected value is over all possible update vectors $\Delta_t(\wvec_t(\stepsizevec))$
for the given step-size and assuming the system started with some $\wvec_0$. 
If the dynamics are ergodic, $\Delta_t(\wvec_t(\stepsizevec))$ does not depend on the initial $\wvec_0$, and is only driven by the underlying state dynamics and the choice of $\stepsizevec$. The step-size can be seen as a control input for this system, with the goal to maintain a stable dynamical system by minimizing $\| \Delta_t(\wvec_t(\stepsizevec)) \|_2^2$ over time.  
%

We derive an algorithm to estimate $\stepsizevec$ for this dynamical system,
which we call AdaGain: Adaptive Gain for Stability. The algorithm is derived for a generic update $\Delta_t(\wvec_t(\stepsizevec))$ that is differentiable w.r.t. the weights $\wvec_t$; we provide specific examples
for particular updates in the appendix, including for linear TD.

\subsection{Generic algorithm with quadratic-complexity}

We derive the full quadratic-complexity algorithm to start, and then introduce approximations to obtain a linear-complexity algorithm. To minimize \eqref{eq_criteria}, we use stochastic gradient descent, and thus need to compute the gradient of $\| \Delta_t(\wvec_t(\stepsizevec)) \|_2^2$ w.r.t. the step-size $\stepsizevec$. For step-size $\stepsize_i$ as the $i$th element in the vector $\stepsizevec$, and $w_{t,j}$ the $j$-th element in vector $\wvec_t$
{\small
\begin{align*}
\frac{\tfrac{1}{2}\partial \| \Delta_t(\wvec_t(\stepsizevec)) \|_2^2}{\partial \stepsize_i}
&= \Delta_t(\wvec_t(\stepsizevec))^\top \frac{\partial \Delta_t(\wvec_t(\stepsizevec))}{\partial \stepsize_i}\\
&= \Delta_t(\wvec_t(\stepsizevec))^\top \sum_{j}^\nparams \frac{\partial \Delta_t(\wvec_t(\stepsizevec))}{\partial w_{t,j}} \frac{\partial w_{t,j}}{\partial \stepsize_i}
.
\end{align*}
}%

The key, then, is to track how a change in the weights impacts the update and how changes in the step-size impact the weights. The first term can be computed instantaneously on this step. For the second term, however, the impact of the step-size on the weights goes back further to previous updates. We show how to obtain a recursive form for this step-size gradient, $\psivec_{t,i} \defeq  \frac{\partial \wvec_{t}}{\partial \stepsize_i} \in \RR^{\nparams}$.
%
{\small
\begin{align*}
\psivec_{t+1,i} 
&= \frac{\partial (\wvec_{t} + \stepsizevec \circ \Delta_t(\wvec_t(\stepsizevec))) }{\partial \stepsize_i}  \\
&= \psivec_{t,i} + \stepsizevec \elwise \sum_{j}\frac{\partial \Delta_t(\wvec_t(\stepsizevec))}{\partial w_{t,j}} \frac{\partial w_{t,j}}{\partial \stepsize_i} + \veczeros{\Delta_{t,i}(\stepsizevec)} \\
&= (\eye + \diag(\stepsizevec) \Gmat_{t}) \psivec_{t,i} + \veczeros{\Delta_{t,i}(\stepsizevec)}
,
\end{align*}
}%

where $\Gmat_{t,j} \defeq \frac{\partial \Delta_t(\wvec_t(\stepsizevec))}{\partial w_{t,j}} \in \RR^\nparams$, $\Gmat_t \defeq [\Gmat_{t,1}, \ldots, \Gmat_{t,\nparams}] \in \RR^{\nparams \times \nparams}$, and \
Therefore, $\psivec_{t+1,i}$ represents a sum of updates, with a recursive weighting on previous $\psivec_{t,i}$ adjusting the weight of previous updates in the sum.
 
We can approximate the gradient using this recursive relationship, without storing all previous samples.
Though the above updates are exact, we obtain an approximation when implementing such a recursive form in practice. When using $\psivec_{t-1,i}$ computed on the last time step $t-1$, this gradient estimate is in fact w.r.t. the previous step-size $\stepsizevec_{t-2}$, rather than $\stepsizevec_{t-1}$. Because these step-sizes are slowly changing, this gradient still provides a reasonable estimate; however, for many steps into the past, the accumulated gradients in $\psivec_{t,i}$ are likely inaccurate. To improve the approximation, and forget old gradients, we introduce a forgetting parameter $0 < \beta < 1$, which focuses the accumulation of gradients in $\psivec_{t,i}$ to a more recent window.  

The gradient update to the step-size also needs to ensure that the step-sizes remain positive. 
Similarly to IDBD, we use an exponential form for the step-size, where $\stepsize = \exp(\beta)$ and $\beta \in \RR$ is updated with (unconstrained) stochastic gradient descent. Conveniently, as we show in the appendix, we do not need to maintain this auxiliary variable, and can simply directly update $\stepsizevec$.

The resulting generic updates for quadratic-complexity AdaGain, with meta step-size $\metastep$, are
{\small
\begin{align}
\stepsizevec_{t} &= \stepsizevec_{t-1}\elwise \exp\left(- \metastep\stepsizevec_{t-1}\elwise(\Psimat_t^\top \Gmat_t^\top \Delta_t) \right)\label{eq_generic_threshold}\\
\psivec_{t+1,i} 
&= (1-\beta) \psivec_{t,i} + \beta \stepsizevec_t \elwise (\Gmat_{t} \psivec_{t,i}) + \beta \veczeros{\Delta_{t,i}}
\nonumber
\end{align}
}
where the exponential is applied element-wise, $\psivec_{0,i} = \zerovec$, $\stepsizevec_0 = 0.1$ (or some initial value), and $(\Psimat_t)_{:,i} = \psivec_{t,i} \text{ with } \Psimat_t \in \RR^{\nparams \times \nparams}$.
For computational efficiency to avoid matrix-matrix multiplication, the order of multiplication for $\Psimat_t^\top \Gmat_t^\top \Delta_t$ should start from the right, as $\Psimat_t^\top (\Gmat_t^\top \Delta_t)$.  
The key complexity in deriving an AdaGain update, then, is simply in computing the Jacobian $\Gmat_t$; given this, the remainder of the algorithm is fixed. For each update $\Delta_t(\wvec_t(\stepsizevec))$, the Jacobian will be different, but is straightforward to compute. 

\newcommand{\jvec}{\hat{\mathbf{j}}}

\subsection{Generic AdaGain algorithm with linear-complexity}

Maintaining the entire matrix $\Psimat_t$ can be prohibitively expensive. 
As was done in IDBD \citep{sutton1992adapting}, one way to avoid maintaining this matrix is to assume that $\frac{\partial w_{t,j}}{\partial \stepsize_i} = 0$ for $i \neq j$. This heuristic reflects that $\stepsize_i$ is likely to have the largest impact on $w_{t,i}$, and less impact on the other entries in $\wvec_t$. 

The modification above for this heuristic is straightforward, simply by setting entries $(\psivec_{t,i})_j = 0$ for $i \neq j$. This results in the simplification
{\small
\begin{align*}
\psivec_{t+1,i} 
&= \psivec_{t,i} + \stepsizevec \elwise \sum_{j}^\nparams \Gmat_{t,j} (\psivec_{t,i})_j + \veczeros{\Delta_{t,i}(\stepsizevec)}\\
&= \psivec_{t,i} + \stepsizevec \elwise \Gmat_{t,i} (\psivec_{t,i})_i + \veczeros{\Delta_{t,i}(\stepsizevec)}
.
\end{align*}
}
Further, since we will then assume that $(\psivec_{t+1,i})_j = 0$ for $i \neq j$, there is no purpose in computing the full vector $\Gmat_{t,i}  (\psivec_{t,i})_i$. Instead, we only need to compute the $i$th entry, i.e., for $\frac{\partial \Delta_{t,i}(\stepsizevec)}{\partial w_{t,i}}$. 
We can then instead define $\hat{\psi}_{t,i}$ to be a scalar approximating $\frac{\partial w_{t,i}}{\partial \stepsize_i}$, with $\hat{\psivec}_{t}$ the vector of these, and $\jvec_t \defeq \left[\frac{\partial \Delta_{t,1}(\stepsizevec)}{\partial \wvec_{t,1}}, \ldots, \frac{\partial \Delta_{t,k}(\stepsizevec)}{\partial \wvec_{t,k}}\right]$
%
%
to define the recursion as $\hat{\psivec}_{t+1} \defeq  \hat{\psivec}_{t} + \stepsizevec \elwise \jvec_{t} \elwise  \hat{\psivec}_{t} + \Delta_t(\wvec_t(\stepsizevec))$, with $\hat{\psivec}_0 = \zerovec$.
The gradient using this approximation, with off-diagonals zero, is
{\small
\begin{align*}
\frac{\tfrac{1}{2}\partial \| \Delta_t(\wvec_t(\stepsizevec)) \|_2^2}{\partial \stepsize_i}
&= \Delta_t(\wvec_t(\stepsizevec))^\top \sum_{j}^\nparams \frac{\partial \Delta_t(\wvec_t(\stepsizevec))}{\partial w_{t,j}} \frac{\partial w_{t,j}}{\partial \stepsize_i}\\
&\approx \Delta_t(\wvec_t(\stepsizevec))^\top \frac{\partial \Delta_t(\wvec_t(\stepsizevec))}{\partial w_{t,i}} \frac{\partial w_{t,i}}{\partial \stepsize_i} \\
&=  \hat{\psi}_{t,i} \Gmat_{t,i}^\top \Delta_t(\wvec_t(\stepsizevec))
\end{align*}
}%

To compute this approximation, for all $i$, we still need to be able to compute $ \Gmat_t^\top \Delta_t(\wvec_t(\stepsizevec))$.
In some cases this is straightforward, as is the case for linear TD (found in the appendix). More generally, we can use R-operators \citep{pearlmutter1994fast} to compute this Jacobian-vector product, or a simple finite difference approximation, as we do in the appendix. Therefore, because we can compute this Jacobian-vector product in linear time, the only approximation is to $\hat{\psivec}_{t}$. 
The update is
\begin{align}
\stepsizevec_{t} &= \stepsizevec_{t-1} \exp\left(- \metastep \ \stepsizevec_{t-1}\elwise\hat{\psivec}_t \elwise ( \Gmat_{t}^\top \Delta_t)  \right) \label{eq_generic_linear_better}\\
\hat{\psivec}_{t+1} 
&=  (1-\beta) \hat{\psivec}_{t} + \beta \stepsizevec_t \elwise \jvec_{t} \elwise  \hat{\psivec}_{t} + \beta \Delta_{t} \nonumber
.
\end{align}

These approximations parallel diagonal approximations, for second-order techniques, which similarly assume off-diagonal elements are zero. Further, $\Gmat_t$ itself is a gradient of the update w.r.t. the weights, where this update was already likely the gradient of the loss w.r.t. the weights. This $\Gmat_t$, therefore, contains similar information as the Hessian. The AdaGain update, therefore, contains some information about curvature, but allows for updates that are not necessarily (true) gradient updates. 

This AdaGain update is generic, but does require computing the Jacobian of a given update, which could be onerous in certain settings. We provide an update, based on finite differences in the appendix, that only requires differences between updates, that we have found works well in practice. 

\section{Experiments in synthetic tasks}
We conduct experiments in several simulation domains to highlight the performance characteristics of meta-descent and quasi-second order methods. In our first experiment we investigate AdaGain and several meta-descent and quasi-second order approaches on a notoriously difficult stationary optimization task. Next we return to the simple state-less tracking problem described in the introduction, and investigate the parameter sensitivity of each method. Our third experiment investigates how different optimization algorithms can stabilize the iterates in sequential off-policy learning problems, which cause SGD-based methods to diverge. We conclude with a comparison of AdaGain and AMSGrad (the best performing quasi-second order method in the first three experiments) for online prediction on data generated by a mobile robot. 

In all the experiments, we use AdaGain layered on-top of an RMSProp update, rather than a vanilla SGD update. As motivated earlier, meta-descent methods are not robust on difficult optimization surfaces, such as with flat or sharp regions. AdaGain provides a practical method to pursue meta-descent strategies that are robust to such realistic optimization problems. We motivate the importance of this choice in our first experiment on a difficult optimization task. 
%
\newcommand{\thetai}{{\theta(t)}}
\newcommand{\mui}{\mu_{t}}
\begin{figure}[h!]
\centering
 \includegraphics[width=0.75\columnwidth]{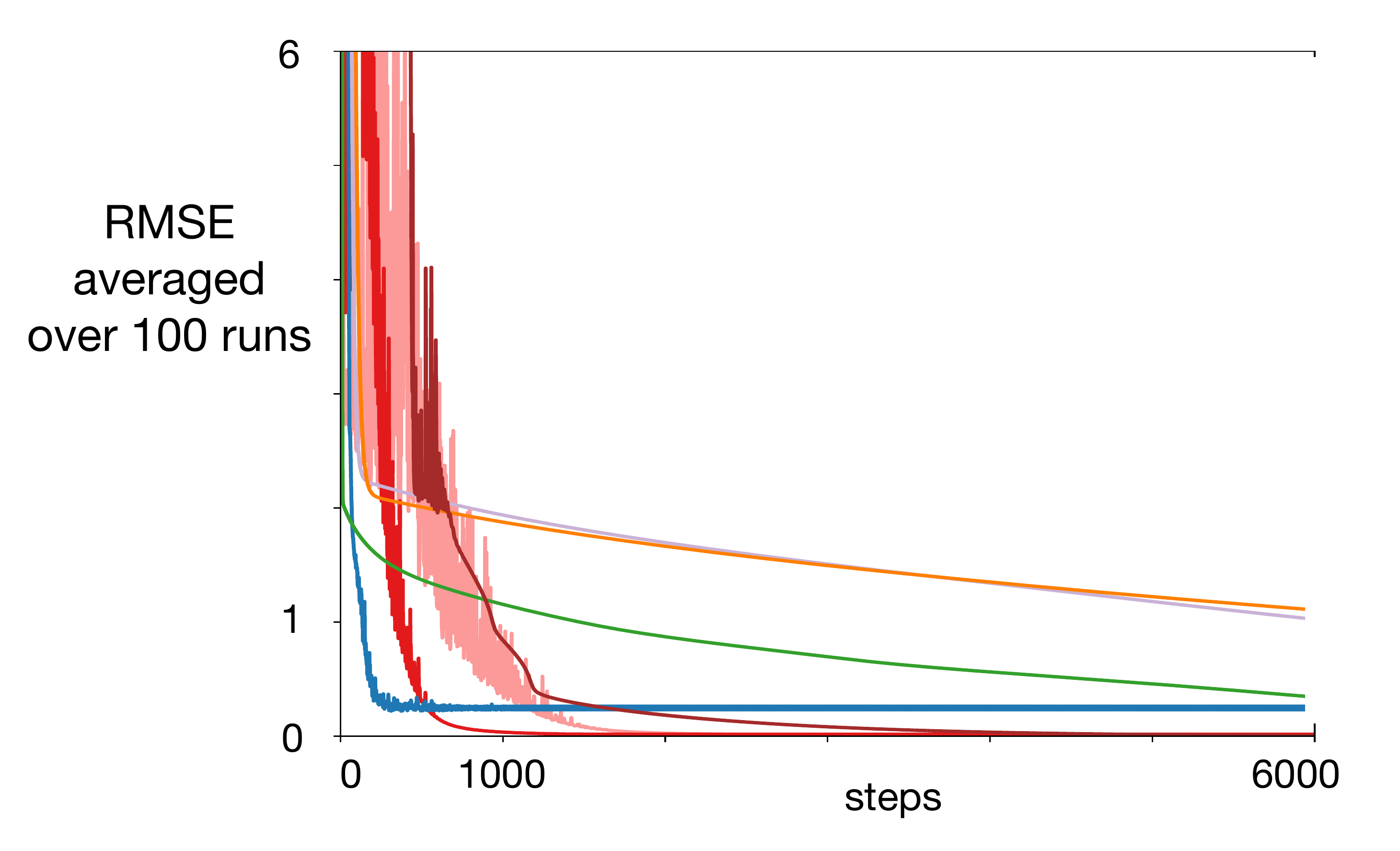}
 \includegraphics[width=0.99\columnwidth]{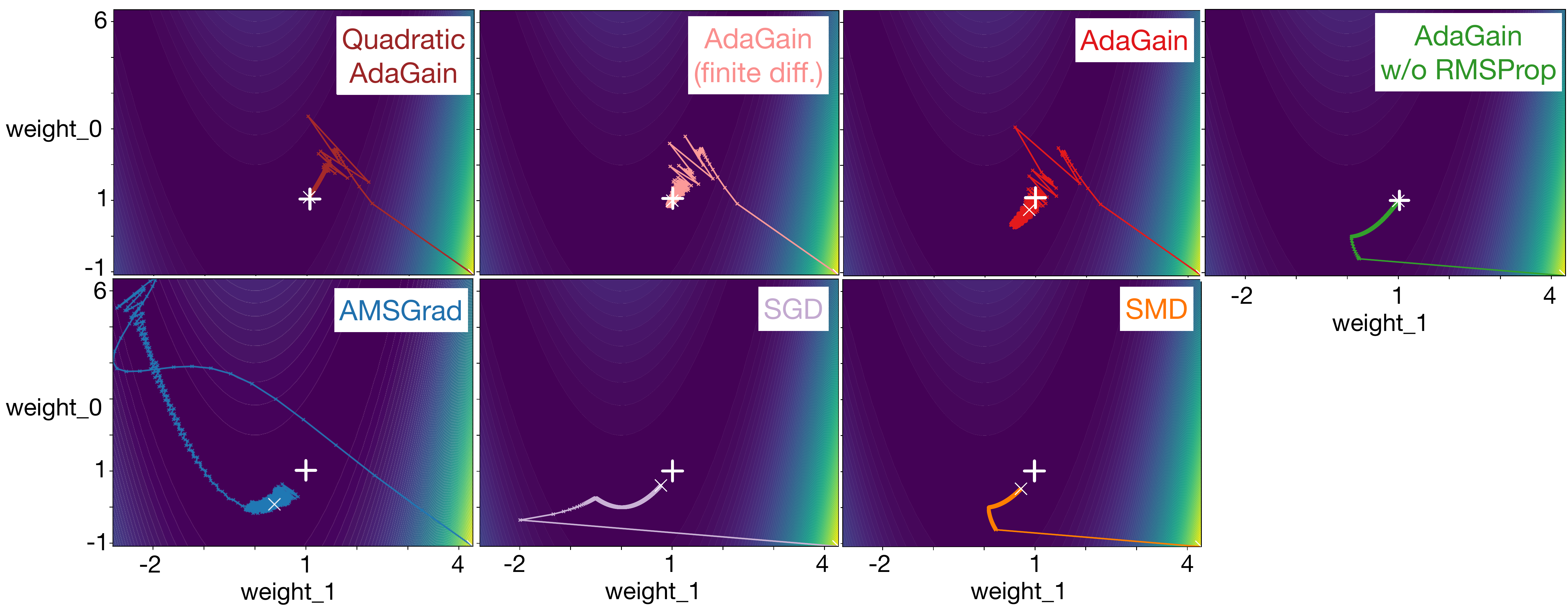}
  \caption{Optimization paths of a single run (with tuned meta-parameters) for several algorithms on the Rosenbrock function. The white $\times$ symbol indicates where in the input space the algorithm converged. The paths represent how each algorithm changes the weights while searching for the minimum. The white $+$ symbol indicates the optimal value for the weights---if $\times$ and $+$ symbol overlap the algorithm has reached the global minimum of the function. Although SGD and SMD appear to quickly approach the minimum, the valley is in fact easy to find, but reaching the $+$ is difficult. Neither method achieves a low final value, and converge slowly. The AdaGain algorithms with RMSProp---including full quadratic AdaGain algorithm, AdaGain with the linear approximation and AdaGain with the linear approximation and finite differences---outperform the other methods in this problem. The finite differences AdaGain algorithm is a generic strategy, that does not require knowledge of the Jacobian, and so can be easily applied to any updates (provided in the appendix). This result highlights that there is not a significant loss in using this approximation, over AdaGain with analytic Jacobians. 
 AdaGain without RMSProp, on the other hand, converges much more slowly, though interestingly it does still outperform SMD. Note although the run above of AdaGain without RMSProp did reach the minimum, that was not true in general as reflected by the learning curve.}\label{fig:rosenbrock}
\end{figure}

\myparagraph{Function optimization.}
The aim of our first experiment is to investigate how AdaGain performs on optimization problems designed to be difficult for gradient descent. 
The Rosenbrock function is a two dimensional non-convex function, and the minimum is inside a flat parabolic shaped valley.  We compared AMSGrad, SGD, and SMD, in each case extensively searching the meta-parameters of each method, averaging performance over 100 runs and 6000 optimization steps. The results are summarized in Figure \ref{fig:rosenbrock}, with trajectory plots of a single run of each algorithm, and the learning curves for all methods. 
AdaGain both learns faster and gets closer to the global optimum than all other methods considered. 
Further, two meta-descent methods, SMD and AdaGain without RMSProp perform poorly. 
This result highlights issues with applying meta-descent approaches without considering the optimization surface, and the importance of having an algorithm like AdaGain which can be combined with quasi-second order methods. 
\begin{figure}[h!]
\centering
  \includegraphics[width=0.4\textwidth]{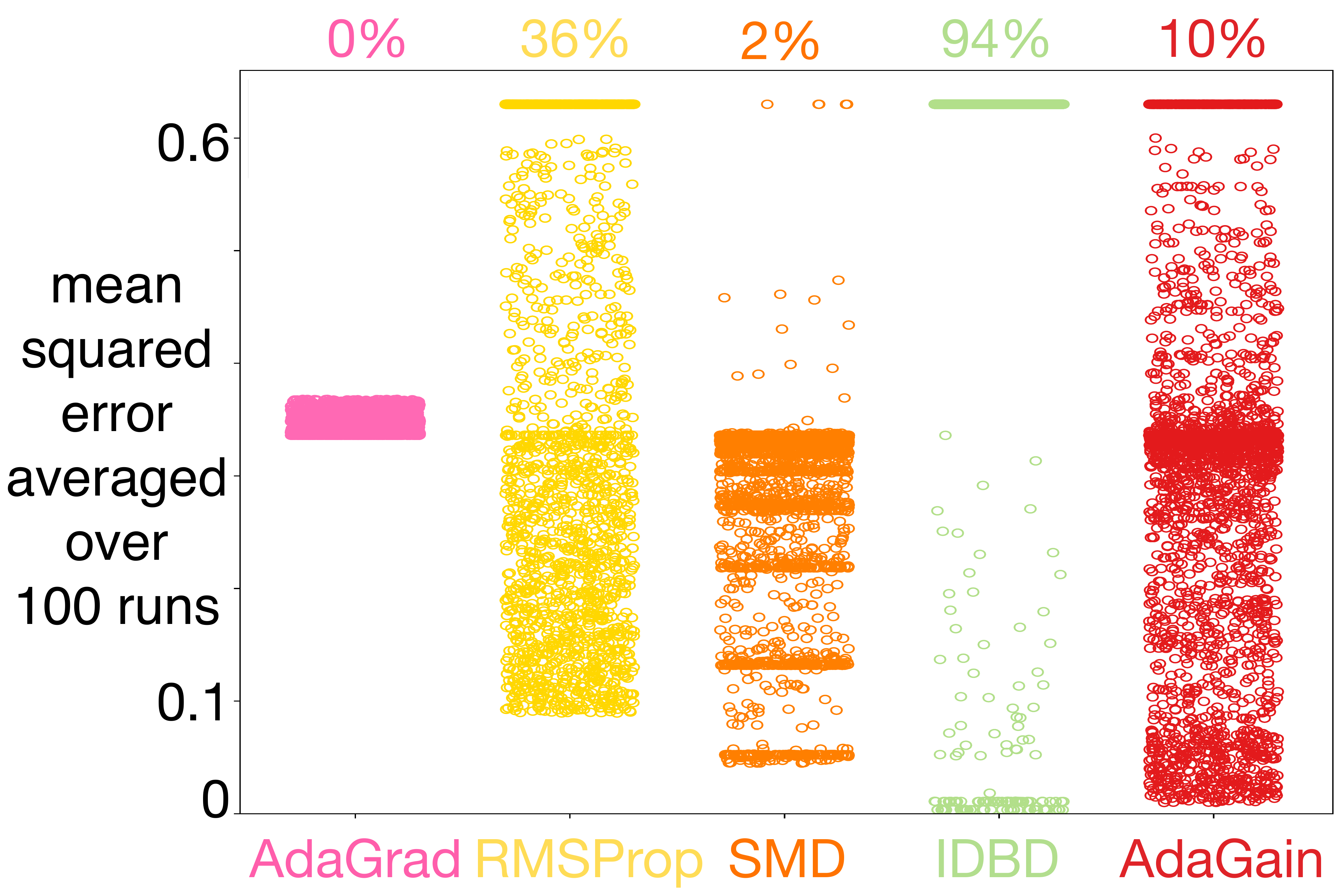}
  \caption{Parameter sensitivity plot for the first 500,000 steps of the stateless tracking problem.  Each circle denotes the average MSE for a single parameter combination of an algorithm. The parameter combinations and respective performance are grouped in vertical columns for each method. The circles in each column are randomly offset within the column horizontally as many parameter settings may achieve almost identical MSE. Circles near the bottom of the plot represent low MSE. Circles arranged in a line in the top-most part of the plot are parameter combinations that either diverged or exceeded a minimum performance threshold, with the percentage of such parameter combinations given in the graph. }\label{fig:gain_water}
\end{figure}

\myparagraph{Stateless tracking problem.}
Recall from Figure \ref{fig:gain_intro}, that several methods performed well in the stateless tracking problem; sensitivity to parameter settings, however, is also important. To help better understand these methods, we constructed a parameter sensitivity graph (Figure \ref{fig:gain_water}). IDBD can outperform AdaGain on this problem (lower MSE), but only a tiny fraction of IDBD's parameter settings achieve good performance. None of AdaGrad's parameter combinations exceeded the threshold, but all combinations resulted in high error compared with AdaGain. Many of the parameter combinations allowed AdaGain to achieve low error, suggesting AdaGain with a simple manual parameter tuning is likely to achieve good performance on this problem, while IDBD likely requires a comprehensive parameter sweep.

\myparagraph{Baird's counterexample.}
Our final synthetic-domain experiment tests the stability of AdaGain's update when combined with the TD($\lambda$) algorithm for off-policy state-value prediction in a Markov Decision Process. We use Baird's counterexample, which causes the weight's learned by off-policy TD($\lambda$)~\citep{SuttonRLIntro} to diverge if a global step-size parameter is used (decaying or otherwise) \citep{baird1995residual,SuttonRLIntro,maei2011gradient}. The key challenge is the feature representation, and the difference between the target and behavior policies. There is a shared redundant feature, and the weight associated seventh feature is initialized to a high value. The target policy always chooses to go to state seven and stay there forever. The behavior policy, on the other hand, only visits state seven 1/7 the time, causing large importance sampling corrections. 

\begin{figure}[h]
    \includegraphics[width=0.99\columnwidth]{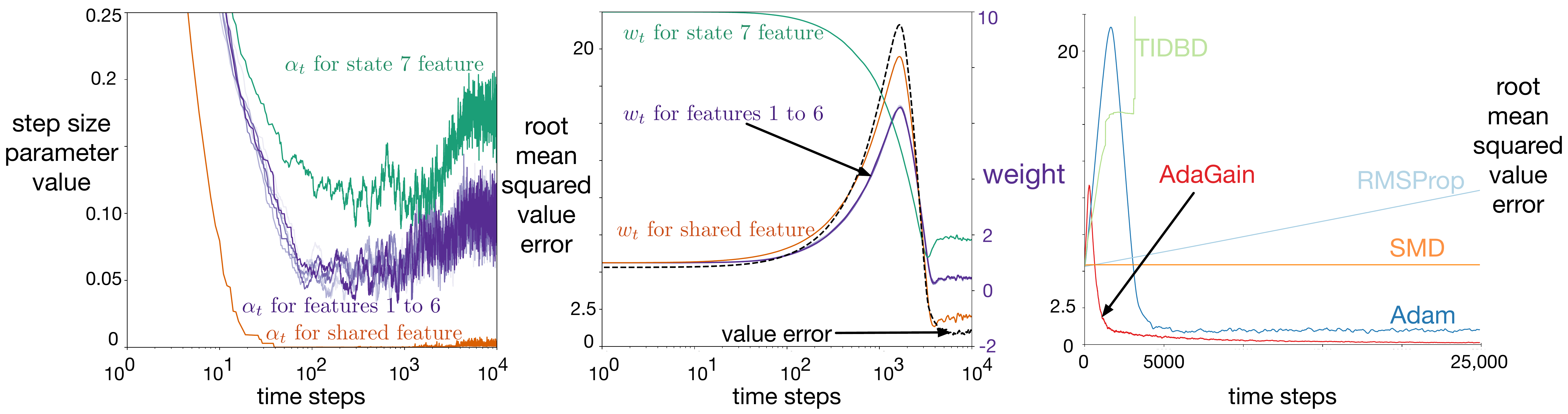}
    \caption{The step-size parameter values over time, and the corresponding weights learned by Adam, and learning curves for several methods in Baird's counterexample. Results averaged over 1000 independent runs. TD($\lambda$) combined with AdaGain achieves the best performance. Adam also prevents divergence, but converges to worse value error. }
    \label{fig:baird_rest}
\end{figure}

We applied AdaGain, Adam, RMSprop, SMD, and TIDBD\citep{kearney2018tidbd}---a
recent extension of the IDBD algorithm --- to adapt the step-sizes of linear TD($\lambda$) on Baird's counterexample. As before, the meta-parameters were extensively swept and the best performing parameters were used to generate the results for comparison. Figure \ref{fig:baird_rest} shows the learning curves of each method. Only AdaGain and Adam are able to prevent divergence. SMD's performance is typical of Baird's counterexample: the meta-parameter search simply found parameters that caused extremely slow divergence. AdaGain learns significantly faster than Adam, and achieves lower error. 

\begin{figure}[h]
    \centering
    \includegraphics[width=0.99\columnwidth]{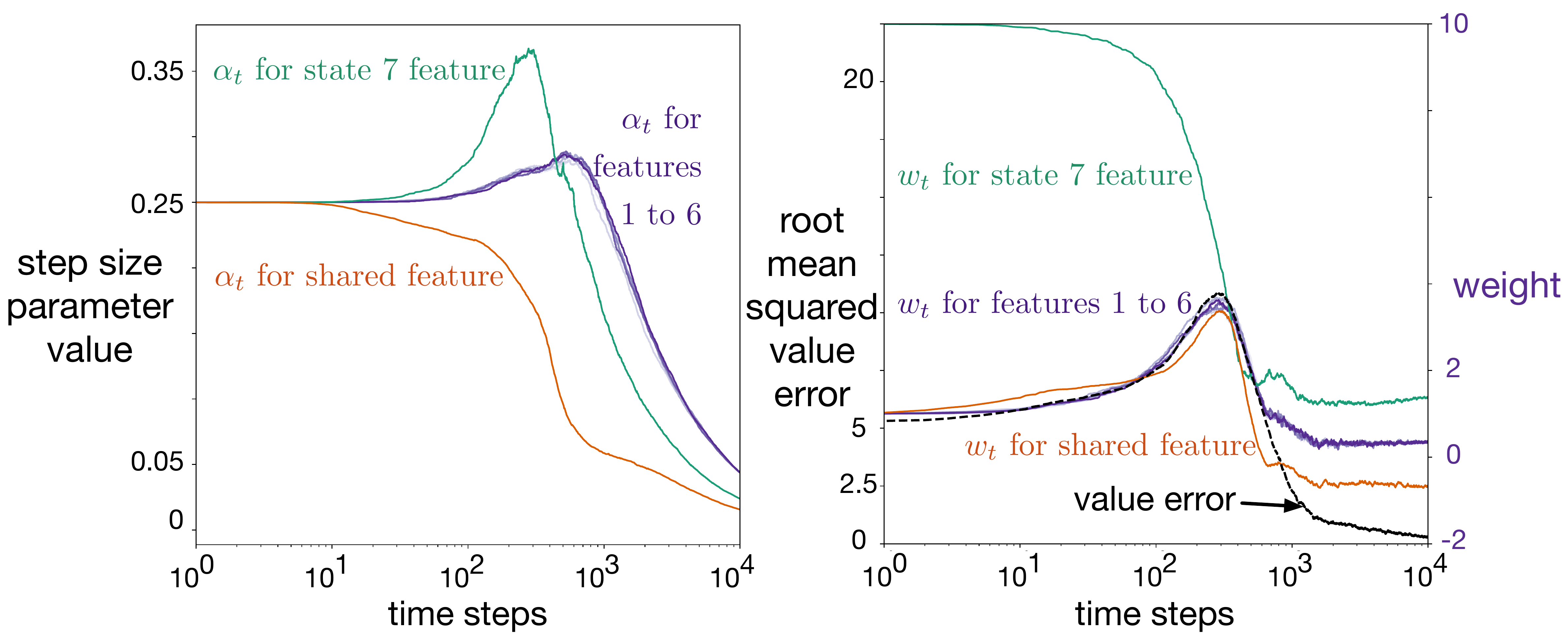}
    \caption{The step-size parameter values over time, and the corresponding weights learned by AdaGain in Baird's counterexample, with results averaged over 1000 independent runs. AdaGain is able to adapt the step-sizes of each feature in such a way that off-policy TD($\lambda$) converges.}
    \label{fig:baird_gain}
\end{figure}

To understand how AdaGain prevents divergence consider Figure \ref{fig:baird_gain}. The left graph shows the step-size values as they evolve over time, and the right graph shows the corresponding weights. Recall, the weight for feature seven is initialized to a high value. AdaGain initially increases feature seven's step-size causing weight seven to quickly fall. In parallel AdaGain reduces the step-size for the redundant feature, preventing incorrect generalization. Over time the weights converge to one of many valid solutions, and the value error, plotted in black on the right side converges to zero. The left plots of Figure \ref{fig:baird_rest} show the same evolution of the weights and step-sizes for Adam. Adam is successful in reducing the step-size for the redundant feature, however the step-sizes of the other features decay quickly and then begin growing again preventing convergence to low value error.           

\begin{figure}[ht]
    \centering
    \includegraphics[width=0.99\columnwidth]{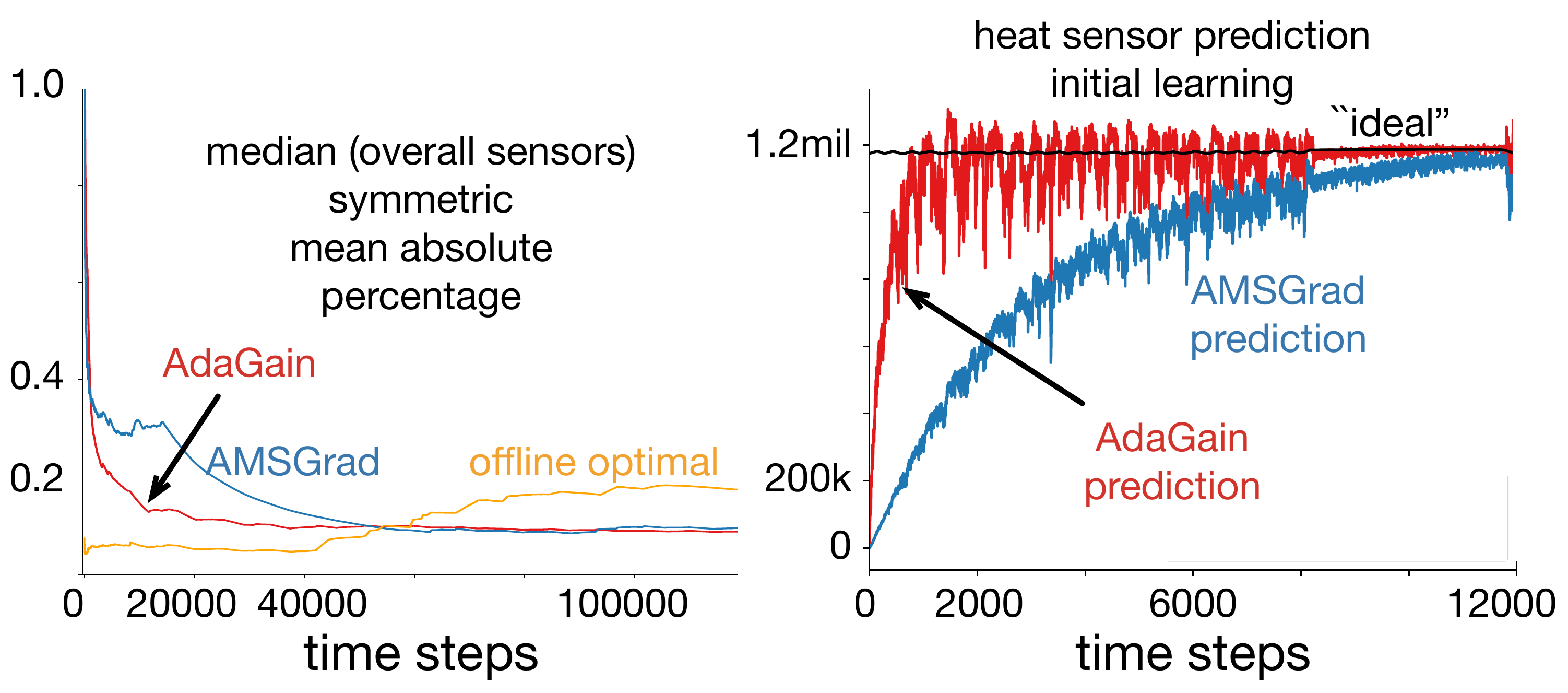}
    \caption{The median symmetric mean absolute percentage error (SMAPE) across all 53 sensors (left), with a plot of the predictions for the heat sensor versus the ideal prediction in early learning (right). 
    The ideal predictions are computed offline using all future data (as described in \citep{modayil2014multi}), but the predictions are learned online and incrementally. The learning curve shows that the predictions learned by AdaGain achieve good accuracy more quickly than those learned by AMSGrad. The right plot highlights early learning performance on the heat sensor---from time zero---illustrating that AdaGain's prediction more quickly approaches the desired magnitude and then maintains good stability. This is particularly notable because the heat sensor targets in this case are unnormalized, obtaining values over 1 million. 
    We also include the optimal predictions computed by solving a system of
    equations offline (again as in \cite{modayil2014multi}). The optimal solution makes use of only the first 40,000 data points for each sensor, reflecting the realistic scenario of computing predictions from a limited batch of data, and later using the offline solution for online prediction. As to be expected the SMAPE for these offline optimal predictions is low on the training data (first 40,000 time steps), and much higher on later data.
  } 
    \label{fig:linearNexting}
\end{figure}

\section{Experiments on robot data}

In our final experiment we recreate nexting \citep{modayil2014multi}, using TD($\lambda$) to make dozens of predictions about the future values of robot sensor readings. We formulate each prediction as estimating the discounted sum of future sensor readings, treating each sensor as a reward signal with discount factor of $\gamma=0.9875$ corresponding to approximately 80 second predictions. Using the freely available nexting data set (144,000 samples, corresponding to 3.4 hours of runtime on the robot), we incrementally processed the data on each step constructing a feature vector from the sensor vector, and making one prediction for each sensor. At the end of learning we computed the "ideal" prediction offline and computed the symmetric mean absolute percentage error of each prediction, and aggregated the 50 learning curves using the median.
We used the same non-linear coarse recoding of the sensor inputs described in the original work, giving 6065 binary feature components for use as a linear representation.

For this experiment we reduced the number of algorithms, using AMSGrad as the best performing quasi-second order method based on our synthetic task experiments and AdaGain as the representative meta-descent algorithm. The meta step-size was optimized for both algorithms.

\begin{figure}[t!]
    \centering
    \includegraphics[width=0.99\columnwidth]{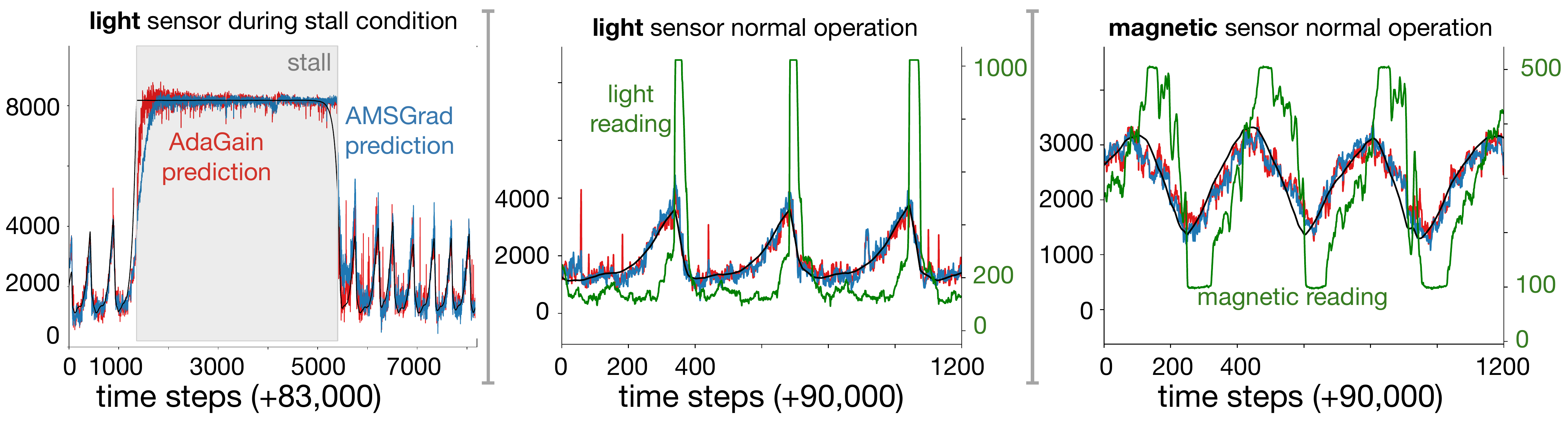}
    \caption{Three snapshots in time of the predictions learned by AdaGain compared with the offline ideal predictions. Each of the three plots highlights a different part of the dataset to give an alternative perspective on the accuracy of AdaGain's learned predictions. The leftmost plot we see a situation where the robot stalled unexpectedly directly in front of a bright light source, saturating the light sensor. Due to this sudden unpredictable event, the predictions of both AdaGain and AMSGrad became incorrect. However, AdaGain more quickly adapts learning to adjust its predictions to reflect the new reality, matching the ideal predictions (black line). Otherwise, these plots show that, in general, AdaGain and AMSGrad can track the ideal prediction similarily.
}
    \label{fig:predsNexting}
\end{figure}

The learning curves in Figure \ref{fig:linearNexting} show a clear advantage for AdaGain in terms of aggregate error over all predictions. Inspecting the predictions of one of the heat sensors reveals why. In early learning, AdaGain much more quickly increases the prediction, to near the ideal prediction, whereas AMSGrad much more slowly reaches this point---over 12000 steps. AdaGain and AMSGrad then both track the the ideal heat prediction similarly, and so obtain similar error for the remainder of learning. This advantage in initial learning is also demonstrated in Figure \ref{fig:predsNexting}, which depicts predictions on two different sensors. For example, AdaGain adapts the predictions more quickly in reaction to the unexpected stall event, but otherwise AdaGain and AMSGrad obtain similar errors. This result also serves as a sanity check for AdaGain, validating that AdaGain does scale to more realistic problems and remains stable in the face of high levels of noise and high-magnitude prediction targets.

\section{Conclusion}

In this work, we proposed a new general meta-descent strategy, to adapt a vector of stepsizes for online, continual prediction problems. 
We defined a new meta-descent objective, that enables a broader class of incremental updates for the base learner, generalizing beyond work specialized to least-mean squares, temporal difference learning and vanilla stochastic gradient descent updates. We derive a recursive update for the stepsizes, and provide a linear-complexity approximation. In a series of experiments, we highlight that meta-descent strategies are not robust to the shape of the optimization surface. The ability to use AdaGain for generic updates enabled us to overcome this issue, by layering AdaGain on RMSProp, a simple quasi-second order approach. We then showed that, with this modification, meta-descent methods can perform better than the more commonly used quasi-second order updates, adapting more quickly in non-stationary tasks. 
\fontsize{9.0pt}{10.0pt} \selectfont
\bibliographystyle{aaai}
\bibliography{paper}

\input{appendix}

\end{document}

%% file: appendix.tex

\appendix

\newcommand{\reg}{R}

\section{Stochastic Meta-Descent algorithm}\label{app_smd}

We recreate the SMD derivation, in our notation, for easier reference.

We compute the gradient of the loss function $\loss(\weights)$, w.r.t. stepsize. 
We derive the full quadratic-complexity algorithm to start, and then introduce approximations to obtain a linear-complexity algorithm. For stepsize $\stepsize_i$ as the $i$th element in the vector $\stepsizevec$, 
\begin{align*}
\frac{\partial \loss(\weights(\stepsizevec)) }{\partial \stepsize_i}
= \sum_{j}^\nparams \frac{\partial \loss(\weights(\stepsizevec))}{\partial w_{t,j}} \frac{\partial w_{t,j}}{\partial \stepsize_i}
\end{align*}

Define the following two vectors, for $w_{t,j}$ the $j$-th element in vector $w_{t,j}$,
\begin{align}
\gvec_{t,j} &\defeq  -\frac{\partial \loss(\weights(\stepsizevec))}{\partial w_{t,j}} \in \RR^{\nparams} \hspace{1.0cm}\text{ the gradient update} \\
\psivec_{t,i} &\defeq  \frac{\partial \wvec_{t}}{\partial \stepsize_i} \in \RR^{\nparams} 
.
\end{align}
We can obtain vector $\psivec_{t,i}$ recursively as
\begin{align*}
\psivec_{t+1,i} 
&= \frac{\partial (\wvec_{t} + \stepsizevec \circ \gvec_{t}) }{\partial \stepsize_i}  
= \frac{\partial \wvec_{t}}{\partial \stepsize_i} + \stepsizevec \circ \frac{\partial \gvec_{t}}{\partial \stepsize_i}  + \veczeros{\gvec_{t,i}(\stepsizevec)} \\
&= \psivec_{t,i} + \stepsizevec \elwise \sum_{j}\frac{\partial  \gvec_{t}}{\partial w_{t,j}} \frac{\partial w_{t,j}}{\partial \stepsize_i} + \veczeros{\gvec_{t,i}(\stepsizevec)}\\
&= \psivec_{t,i} - \stepsizevec \elwise (\Hmat_{t} \psivec_{t,i}) + \veczeros{\gvec_{t,i}(\stepsizevec)}\\
&= (\eye - \diag(\stepsizevec) \Hmat_{t}) \psivec_{t,i} + \veczeros{\gvec_{t,i}(\stepsizevec)}
.
\end{align*}

The resulting generic updates for quadratic-complexity SMD, with meta stepsize $\metastep$, are
\begin{align}
\stepsizevec_{t} &= \stepsizevec_{t-1} \exp\left(\metastep\stepsizevec_t\elwise \Psimat_t^\top \gvec_t \right)\\ 
&\hspace{2.0cm} \text{ for } (\Psimat_t)_{:,i} = \psivec_{t,i} \text{ with } \Psimat_t \in \RR^{\nparams \times \nparams} \label{eq_generic_threshold}\nonumber\\
\Hmat_t &=  \text{Hessian of $\loss_t$ w.r.t. $\weights_t$}.  \nonumber\\
\psivec_{t+1,i} 
&= (1-\beta) \psivec_{t,i} - \beta \stepsizevec_t \elwise (\Hmat_{t} \psivec_{t,i}) + \beta \veczeros{\gvec_{t,i}}
\nonumber
\end{align}
$\psivec_{0,i} = \zerovec$ and $\stepsizevec_0 = 0.1$ (or some initial value).
As with AdaGain, the Hessian-vector product $\Hmat_{t} \psivec_{t,i}$ can be computed efficiently, using R-operators. Here, it is irrelevant, because we maintain the quadratic $\Psimat$.

For the linear-complexity algorithm, again we set entries $(\psivec_{t,i})_j = 0$ for $i \neq j$. Let $\Hmat_{t,i}$ be the $i$th column of the Hessian. This results in the simplification
\begin{align*}
\psivec_{t+1,i} 
&= \psivec_{t,i} - \stepsizevec \elwise \sum_{j}^\nparams \Hmat_{t,j} (\psivec_{t,i})_j + \veczeros{\gvec_{t,i}(\stepsizevec)}\\
&= \psivec_{t,i} - \stepsizevec \elwise \Hmat_{t,i} (\psivec_{t,i})_i + \veczeros{\gvec_{t,i}(\stepsizevec)}
.
\end{align*}
Further, since we will then assume that $(\psivec_{t+1,i})_j = 0$ for $i \neq j$, there is no purpose in computing the full vector $\Hmat_{t,i}  (\psivec_{t,i})_i$. Instead, we only need to compute the $i$th entry, i.e., for $\frac{\partial \gvec_{t,i}(\stepsizevec)}{\partial w_{t,i}}$. 
We can then instead define $\hat{\psi}_{t,i}$ to be a scalar approximating
$\frac{\partial w_{t,i}}{\partial \stepsize_i}$, with $\hat{\psivec}_{t}$ the
vector of these, and the diagonal of the Hessian
\begin{equation}
\hat{\hvec}_t \defeq \left[\frac{\partial^2 \loss(\weights(\stepsizevec)}{\partial \wvec_{t,1}^2}, \ldots, \frac{\partial^2 \loss(\weights(\stepsizevec)}{\partial \wvec_{t,k}^2}\right]
\end{equation}
to define the recursion as $\hat{\psivec}_{t+1} \defeq  \hat{\psivec}_{t} - \stepsizevec \elwise \hat{\hvec}_{t} \elwise  \hat{\psivec}_{t} + \gvec_{t}(\stepsizevec)$, with $\hat{\psivec}_0 = \zerovec$.
The gradient using this approximation, with off-diagonals zero, is
\begin{align*}
\frac{\partial \loss(\weights(\stepsizevec)) }{\partial \stepsize_i}
&= \sum_{j}^\nparams \frac{\partial \loss(\weights(\stepsizevec))}{\partial w_{t,j}} \frac{\partial w_{t,j}}{\partial \stepsize_i}\\
&\approx \frac{\partial \loss(\weights(\stepsizevec))}{\partial w_{t,i}} \frac{\partial w_{t,i}}{\partial \stepsize_i}\\
&=  \hat{\psi}_{t,i} \gvec_{t,i}
\end{align*}
The resulting update to the stepsize is
\begin{align}
\stepsizevec_{t} &= \stepsizevec_{t-1} \exp\left(\metastep\stepsizevec_t\elwise \hat{\psivec}_t \elwise \gvec_t  \right)\\
\hat{\psivec}_{t+1} 
&=  (1-\beta) \hat{\psivec}_{t} - \beta \stepsizevec_t \elwise \hat{\hvec}_{t} \elwise  \hat{\psivec}_{t} + \beta \gvec_{t} \nonumber
.
\end{align}
%



\textbf{Difference to original SMD algorithm:}
Now, surprisingly, the above algorithm differs from the algorithm given for SMD. But, that derivation appears to have a flaw, where the gradients of weights taken w.r.t. to a vector of stepsizes is assumed to be a vector. Rather, with the same off-diagonal approximation we use, it should be a diagonal matrix, and then they would also only get a diagonal Hessian. For completeness, we include their algorithm, which uses a full Hessian-vector product.
\begin{align}
\stepsizevec_{t} &= \stepsizevec_{t-1} \exp\left(\metastep \stepsizevec_t\elwise \hat{\psivec}_t \elwise \gvec_t  \right)\\
\hat{\psivec}_{t+1} 
&=  \hat{\psivec}_{t} - \stepsizevec_t \elwise \Hmat_t \hat{\psivec}_{t} + \gvec_{t} \nonumber
.
\end{align}
Note that a follow-up paper that tested SMD \citep{wu2018understanding} uses this update, but does not have an error, because they use a \emph{scalar} step size. In fact, in the SMD paper, if the step size had been a scalar, then their derivation would be correct.

\textbf{The addition of $\beta$:}
The original SMD algorithm did not use forgetting with $\beta$. In our experiments, however, we consider SMD with $\beta$---which performs significantly better---since our goal is not to compare directly with SMD, but rather to compare the choice of objectives. 

\section{Derivations for AdaGain updates}\label{app_updates}

Consider again the generic update
%
\begin{align}
\wvec_{t+1} = \wvec_t + \stepsizevec \circ \Delta_t
\end{align}
where $\Delta_t \in \RR^\xdim$ is the update for this step,
for weights $\wvec_t \in \RR^\featdim$ and constant vector stepsize $\stepsizevec$ and the operator $\elwise$ denotes element-wise multiplication. 

\subsection{Maintaining non-negative stepsizes in AdaGain}\label{app_threshold}

One straightforward option to maintain non-negative stepsizes is to define a constraint on the stepsize. We can prevent the stepsize from going below a small threshold $\lowerthreshold$ (e.g., $\lowerthreshold = 0.001$), ensuring
positive stepsizes. The projection onto this constraint set after each gradient descent step simply involves applying the
operator $(\cdot)_{\lowerthreshold}$, which thresholds any values below $\lowerthreshold > 0$ to $\lowerthreshold$. We experimented with this strategy compared to the mentioned exponential form, and found it performed relatively similarly, but required an extra parameter to tune. 

Another option---and the one we use in this work---is to use an exponential form for the stepsize, so that it remains positive.
One form, used also by IDBD, is to use $\stepsizevec = \exp(\betavec)$. 
The algorithm, with or without an exponential form, remains essentially identical to the thresholded version, because
\begin{align*}
\frac{\tfrac{1}{2}\partial \| \Delta_t(\stepsizevec(\betavec)) \|_2^2}{\partial \beta_i}
&= \Delta_t(\stepsizevec(\betavec)) \frac{\Delta_t(\stepsizevec(\betavec))}{\partial \stepsize_i} \frac{\partial \stepsize_i}{\partial \beta_i}
.
\end{align*}
Therefore, we can still recursively estimate the gradient with the same approach, regardless of how the stepsize $\stepsizevec$ is constrained. For the thresholded form, we simply use the gradient $\Delta_t(\stepsizevec(\betavec)) \frac{\Delta_t(\stepsizevec)}{\partial \stepsize_i}$ and then project (i.e., threshold). For the exponential form, the gradient update for $\stepsizevec$ is simply used within an exponential function, as described below.

Consider directly maintaining $\betavec$, which is unconstrained. 
For the function form $\stepsize_i = \exp(\beta_i)$, the partial derivative $\frac{\partial \stepsize_i}{\partial \beta_i}$ is simply equal to $\stepsize_i$
and so the gradient update includes an additional $\stepsize_i$ in front. 
This can more explicitly be maintained, without an additional variable, by noticing that for gradient $g_i  = \stepsize_i \Delta_t(\stepsizevec(\betavec)) \frac{\Delta_t(\stepsizevec(\betavec))}{\partial \stepsize_i}$ for $\beta_{t,i}$
\begin{align*}
\stepsize_{t+1,i} &= \exp(\beta_{t+1,i}) \\
&= \exp(\beta_{t,i} - \metastep g_i)\\
&= \exp(\beta_{t,i}) \exp(-\metastep g_i)\\
&= \stepsize_{t,i} \exp(-\metastep g_i)
\end{align*}
Therefore, we can still directly maintain $\stepsizevec$. The resulting update to $\stepsizevec$ is simply
\begin{equation}
\stepsizevec_{t} = \stepsizevec_{t-1} \exp\left(-\metastep \stepsizevec_t \elwise \hat{\psivec}_t \elwise ( \Gmat_{t}^\top \Delta_t) \right) 
\end{equation}
%

Other multiplicative updates are also possible. \citet{schraudolph1999local} uses an exponential update, but uses an approximation with a maximum, to avoid the expensive computation of the exponential function. \citet{baydin2018online} uses a similar multiplicative update, but without a maximum. 

\subsection{AdaGain for linear TD}\label{app_td}

In this section, we derive $\gvec_t$ for a particular algorithm, namely linear TD. LMS updates can be obtained as special cases, by setting $\gamma = 0$. We then provide a more general update algorithm---which does not require knowledge of the form of the update---in the next section.
One advantage of AdaGain is that it is derived generically, allowing extensions to many online algorithms, unlike IDBD, and variants which are derived specifically for the squared TD-error. 

We first provide the AdaGain updates for linear TD($\lambda$), and then provide the derivation below. 
For TD($\lambda$), the update is
\begin{align}
\delta_t &\defeq r_{t+1} + \gamma_{t+1} \featvec_{t+1}^\top \wvec_{t} -  \featvec_{t}^\top \wvec_{t}\nonumber\\
\Delta_t &\defeq \delta_t \evec_t \nonumber\\ 
\nonumber\\
\stepsizevec_{t} &= \stepsizevec_{t-1}\exp(-\metastep (\Delta_t^\top \evec_t)\stepsizevec_{t-1}\elwise \dvec_t \elwise \hat{\psivec}_t ) \label{eq_td_updates}\\
\hat{\psivec}_{t+1} &= (1-\beta) \hat{\psivec}_{t,i} + \beta \stepsizevec_t \elwise \evec_{t} \elwise \dvec_{t} \elwise \hat{\psivec}_{t} + \beta \Delta_{t} \nonumber
\end{align}
where $\stepsizevec_0 = 0.1$, $\hat{\psivec}_0 = \zerovec$, $\gamma_t \defeq \gamma(S_t, A_t, S_{t+1})$ 

To derive the update for $\stepsizevec$, we need to compute the gradients of the updates, particularly $ (\gvec_t)_i = \frac{\partial \Delta_{t,i}}{\partial \wvec_{t,i}}$ or for the full algorithm, the Jacobian $\Gmat$.  
\begin{align*}
\frac{\partial \Delta_t}{\partial \wvec_{t,i}}
&= \evec_t \frac{\partial \delta_t}{\partial \wvec_{t,i}}\\ 
 &=  \evec_t \frac{\partial }{\partial \wvec_{t,i}} (r_{t+1} + \gamma_{t+1} \featvec_{t+1}^\top \wvec_t -  \featvec_{t}^\top \wvec_t ) \\
  &=  \evec_t \left(\gamma_{t+1} \featvec_{t+1} -  \featvec_{t}\right)_i 
.
\end{align*}
Letting $\dvec_t\defeq\gamma_{t+1} \featvec_{t+1} -  \featvec_{t}$, the Jacobian is $\Gmat_t = \evec_t \dvec_t^\top$ and the diagonal approximation is $\gvec_t = \evec_t \elwise \dvec_t$. Because of the form of the Jacobian, we can actually use it in the update to $\stepsizevec$, though not in computing $\hat{\psivec}_t$, if we want to maintain linearity. 
The quadratic complexity algorithm uses $\Gmat$ as given
\begin{align*}
  \stepsizevec_{t} &= \stepsizevec_{t-1} \exp(-\metastep (\Delta_t^\top\dvec_t)\stepsizevec_{t-1}\elwise(\Psimat_t^\top\evec_t)) \\
\psivec_{t,i} 
&= (1-\beta) \psivec_{t-1,i}\\ &+ \beta \stepsizevec \elwise (\evec_{t-1} \dvec_{t-1}^\top \psivec_{t-1,i}) + \beta \veczeros{\Delta_{t-1,i}}
\nonumber
\end{align*}
The linear complexity algorithm uses $\gvec_t$ to update $\hat{\psivec}_t$, giving the stepsize update in \eqref{eq_td_updates}
\begin{align*}
\stepsizevec_{t} &= \stepsizevec_{t-1}\exp(-\metastep (\Delta_t^\top \dvec_t)\stepsizevec_{t-1}\elwise \evec_t \elwise \hat{\psivec}_t ) \\
\hat{\psivec}_{t+1} 
&= (1-\beta) \hat{\psivec}_{t,i} + \beta \stepsizevec_t \elwise \evec_{t} \elwise \dvec_{t} \elwise \hat{\psivec}_{t} + \beta \Delta_{t}
\nonumber
\end{align*}

\subsection{Generic AdaGain algorithm}\label{app_generic}

To avoid requiring knowledge about the algorithm update and its derivatives, we can provide an approximation
to the Jacobian-vector product and the diagonal of the Jacobian, using finite differences. As long as the update function
for the algorithm can be queried multiple times, this algorithm can be easily applied to any update. 

To compute the Jacobian-vector product, we use the fact that this corresponds to a directional derivative. Notice that $\Gmat_t^\top \Delta_t$ corresponds to the vector of directional derivatives for each component (function) in the update $\Delta_t$, in the direction of $\uvec = \Delta_t$, because the dot-product separates in $\Gmat_{t,1}^\top \uvec, \ldots, \Gmat_{t,\nparams}^\top \uvec$. Therefore, for update function $\Delta: \RR^\nparams \rightarrow \RR^\nparams$ (such as the gradient of the loss), we get for small $r = 0.001$, 
\begin{equation}
\Gmat_t^\top \Delta_t \approx \frac{\Delta(\wvec + r \uvec) - \Delta(\wvec - r \uvec)}{2r}
\end{equation}
For the diagonal of the Jacobian, we can again use finite differences. An efficient finite difference computation is proposed within the 
simultaneous perturbation stochastic approximation algorithm \citep{spall1992spsa}, which uses a random perturbation vector $\epsilonvec$ to compute the centered difference $\frac{(\Delta(\wvec + r \epsilonvec) - \Delta(\wvec - r \epsilonvec))_i}{2 r \epsilonvec_i}$. This formula provides an approximation to the gradient of the $i$ entry in the update $\Delta_t$ with respect to weight $i$; when computed for all $i$, this approximates the diagonal of the Jacobian $\jvec_{t}$.
To avoid additional computation, we can re-use the above difference with perturbation $\uvec$, rather than a random vector $\epsilonvec$. To avoid division by zero, if $\uvec$ contains a zero entry, we threshold the normalization with a small constant $10^{-6}$ to give
\begin{equation}
\jvec_{t} \approx \frac{\Delta(\wvec + r \uvec) - \Delta(\wvec - r \uvec))}{2r} \elwise (1/\sign(\uvec) \max(10^{-6}, |\uvec|))
\end{equation}
where division is element-wise. 
another approach would be to sample a random direction $\epsilonvec$ for this finite difference and use $\Delta(\wvec + \epsilonvec) - \Delta(\wvec)$, divided by the absolute value of each element of $\epsilonvec$.
We found empirically that using the same direction as $\Delta_t$ was actually more effective, and more computationally efficient, so we propose that approach.

Using these approximations, we can compute the update to the stepsize as in Equation \eqref{eq_generic_linear_better}, repeated here for easy reference
\begin{align*}
\stepsizevec_{t} &= \stepsizevec_{t-1} \exp\left(- \metastep \stepsizevec_{t-1}\elwise \hat{\psivec}_t \elwise ( \Gmat_{t}^\top \Delta_t)  \right) \\
\hat{\psivec}_{t+1} 
&=  (1-\beta) \hat{\psivec}_{t} + \beta \stepsizevec_t \elwise \jvec_{t} \elwise  \hat{\psivec}_{t} + \beta \Delta_{t} 
.
\end{align*}

\begin{figure}[h!]
\centering
\includegraphics[width=0.4\textwidth]{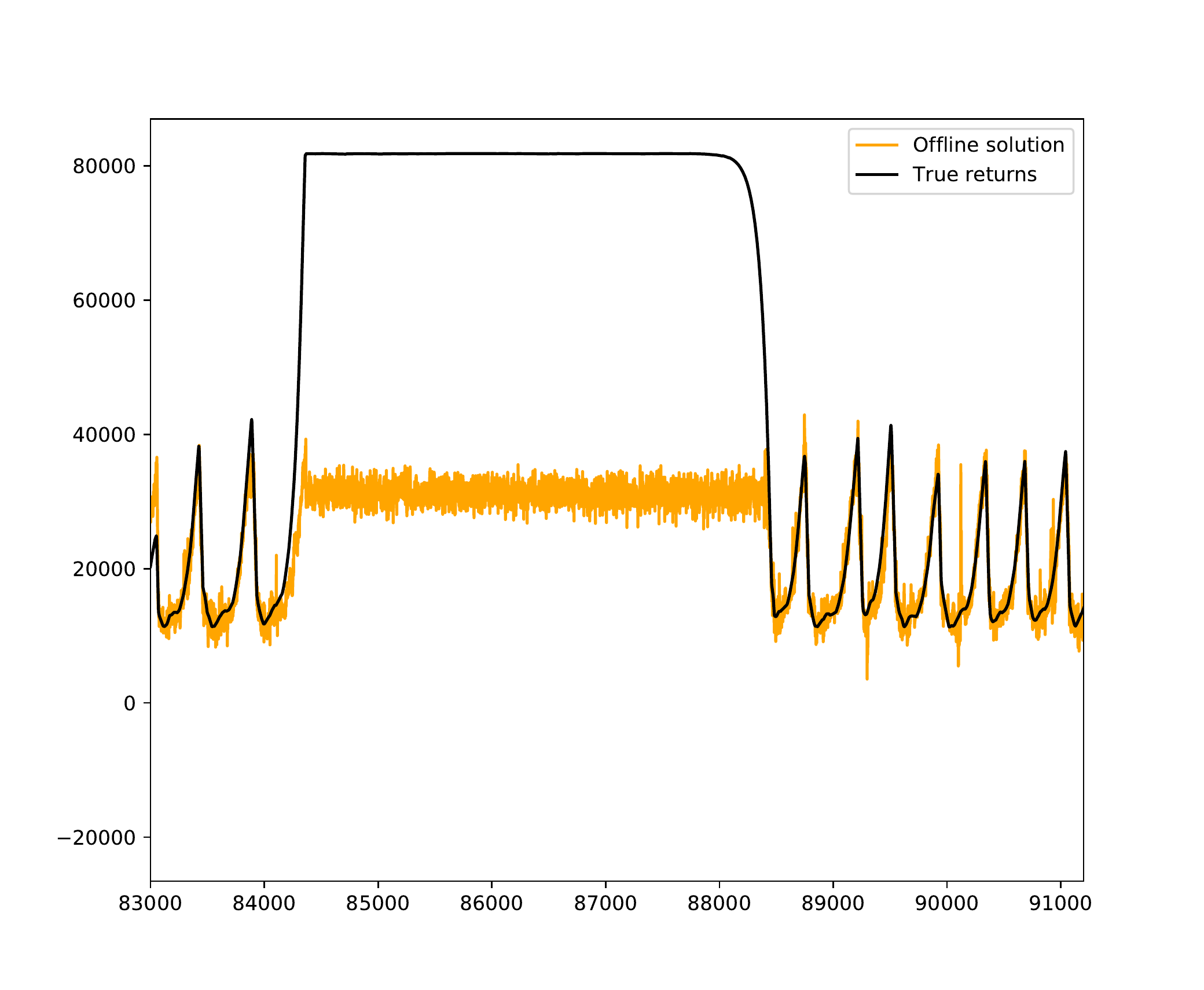}
\includegraphics[width=0.4\textwidth]{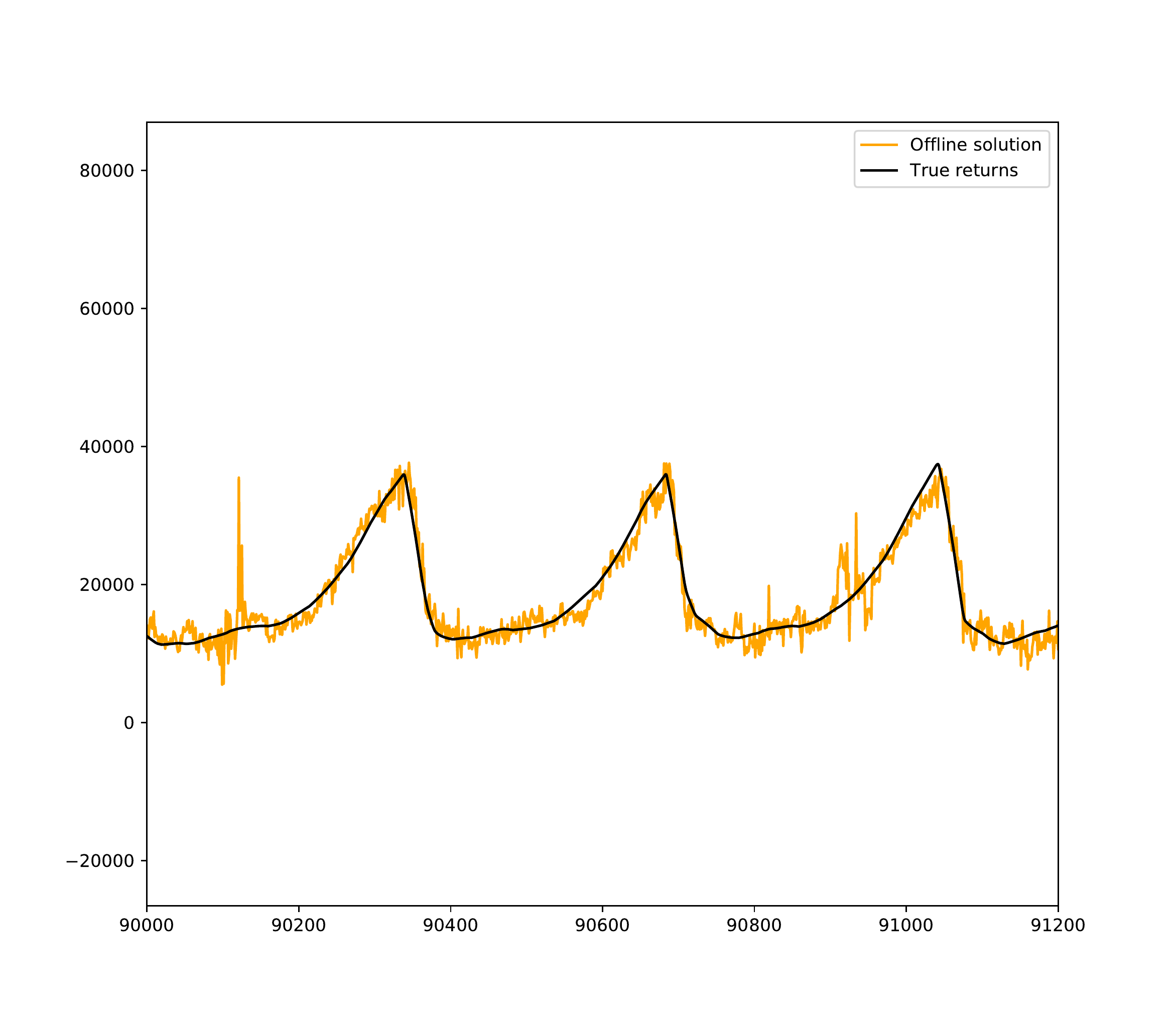}
\caption{Depicted above are the offline optimal predictions during the light sensor stall, and during the light sensor's normal operation (see Figure \ref{fig:predsNexting}). The optimal offline solution was trained by computing the linear least-squares solution for the first 40,000 data points, and using that solution to make predictions on the rest of the dataset.}

\end{figure}